\def\year{2022}\relax
\documentclass[letterpaper]{article} 
\usepackage{aaai22}  
\usepackage{times}  
\usepackage{helvet}  
\usepackage{courier}  
\usepackage[hyphens]{url}  
\usepackage{graphicx} 
\usepackage{natbib}  
\usepackage{caption} 
\DeclareCaptionStyle{ruled}{labelfont=normalfont,labelsep=colon,strut=off} 
\usepackage{algorithm}
\usepackage{algorithmic}

\usepackage{newfloat}
\usepackage{listings}
\floatstyle{ruled}
\newfloat{listing}{tb}{lst}{}
\floatname{listing}{Listing}
\usepackage{soul,xcolor}
\usepackage{multirow}
\usepackage{subfig}
\usepackage{enumitem}
\usepackage{array}
\newcolumntype{C}[1]{>{\centering\arraybackslash\hspace{0pt}}p{#1}}
\usepackage{lipsum,graphicx}
\usepackage{amsmath,amssymb}

\usepackage{colortbl}
\usepackage{arydshln}
\definecolor{LightCyan}{rgb}{0.88,1,1}

\usepackage{array}
\newcolumntype{H}{>{\setbox0=\hbox\bgroup}c<{\egroup}@{}}
 \usepackage{MnSymbol,wasysym}
 
\newcommand{\name}{\textsf{MORE}}

\newcommand{\modelname}{\textsf{ExMore}}

\title{{\em Nice perfume. How long did you marinate in it? \smiley{}\\} Multimodal Sarcasm Explanation}
\author{
    Poorav Desai, Tanmoy Chakraborty, Md Shad Akhtar
}
\affiliations{
    Indraprastha Institute of Information Technology, Delhi (IIIT Delhi), India\\
    \{desai19010@iiitd.ac.in, tanmoy, shad.akhtar\}@iiitd.ac.in
}

\begin{document}

\maketitle

\begin{abstract}
Sarcasm is a pervading linguistic phenomenon and highly challenging to explain due to its subjectivity, lack of context and deeply-felt opinion. In the multimodal setup, sarcasm is conveyed through the {\em incongruity between the text and visual entities}. Although recent approaches deal with sarcasm as a classification problem, it is unclear why an online post is identified as sarcastic. Without proper explanation, end users may not be able to perceive the underlying sense of irony. In this paper, we propose a novel problem -- \textbf{\underline{Mu}ltimodal \underline{S}arcasm \underline{E}xplanation} (MuSE) -- given a multimodal sarcastic post containing an image and a caption, we aim to {\em generate a natural language explanation} to reveal the intended sarcasm. To this end, we develop \name, a new dataset with explanation of $3510$ sarcastic multimodal posts. Each explanation is a natural language (English) sentence describing the hidden irony. We benchmark \name\ by employing a multimodal Transformer-based architecture. It incorporates a cross-modal attention in the Transformer's encoder which attends to the distinguishing features between the two modalities. Subsequently, a BART-based auto-regressive decoder is used as the generator. Empirical results demonstrate convincing results over various baselines (adopted for MuSE) across five evaluation metrics. We also conduct human evaluation on predictions and obtain Fleiss' Kappa score of $0.4$ as a fair agreement among 25 evaluators.   
\end{abstract}

\section{Introduction}
Sarcasm\footnote{https://www.merriam-webster.com/dictionary/sarcasm} refers to the use of satirical or ironic statements usually to hurt, insult, or offend someone. The surface meaning of such statements is usually different from the intended meaning, and to comprehend the sarcasm, one needs to be aware of the context in which the statement was uttered. \citet{joshi-2015-harnessing-context-incongruity} suggested the presence of incongruity as a vital signal for sarcasm. Traditionally, the research around sarcasm analysis revolves around the detection of underlying sarcasm in text \cite{campbell-2012-sarcasm,riloff-etal-2013-sarcasm}. In recent years, the exploitation of multimodal signals \textit{viz.} image, video or audio, is on the rise for detecting sarcasm \cite{schifanella2016detecting, castro-etal-2019-towards, sangwan:multimodal:sarcasm:ijcnn:2020}. With multimodal signals, the scope of incongruity in sarcastic posts expands to inter-modality  and intra-modality incongruity. Most  existing systems rely on the interaction among the modality-specific latent representations for leveraging incongruity. For example, \citet{sangwan:multimodal:sarcasm:ijcnn:2020} employed a gating mechanism to fuse the two modalities.            
\begin{figure}[!t]
    \centering
    \subfloat[Difference of MuSE with non-sarcastic interpretation\label{fig:example:difference}]{
    \resizebox{\columnwidth}{!}{
    \begin{tabular}{ccc}
         \includegraphics[width=.4\columnwidth]{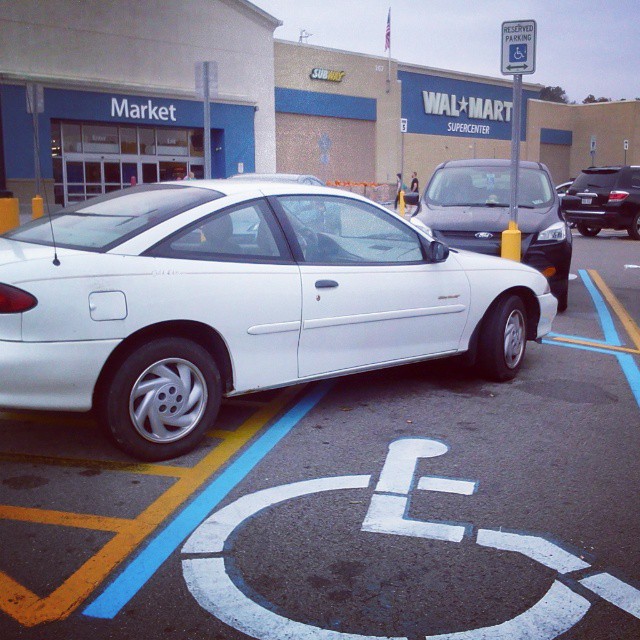}\\
         \multirow{1}{25em}{{\bf Caption:} This guy gets a gold star for excellent parking.} \\ \hline
        \rowcolor{LightCyan} \\ 
        \rowcolor{LightCyan} \multirow{-2}{25em}{{\bf Explanation:} this guy has parked his car partially covering the parking slot for handicapped.} \\ 
         \hline
         \multirow{2}{25em}{{\bf Non-Sarcastic utterance:} This guy does not get a gold star for bad parking.} \\ \\
    \end{tabular}}
    }

    \subfloat[Similarity of MuSE and non-sarcastic interpretation.\label{fig:example:similarity}]{
    \resizebox{\columnwidth}{!}{
    \begin{tabular}{ccc}
         \includegraphics[width=15em, height=5em]{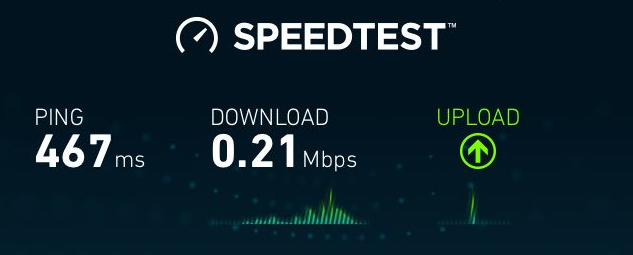}\\
         \multirow{2}{25em}{{\bf Caption:} Internet is awesome. i love my dialup internet! show me the 70mbps!  \#wtf.} \\ \\ \hline
         \rowcolor{LightCyan} \multirow{1}{25em}{{\bf Explanation:} This internet is terrible, I hate dialup internet.} \\ \hline
         \multirow{2}{25em}{{\bf Non-Sarcastic utterance:} This internet is not awesome, I hate dial up internet.} \\ \\ 
    \end{tabular}}
    }
    \caption{Example scenarios showing the multimodal sarcasm explanation task compared to the (textual) non-sarcastic interpretation task \cite{joshi:non-sarcastic:2019}. Non-sarcastic interpretation is primarily negation of caption.}
    \label{fig:example}
    \vspace{-5mm}
\end{figure}

\paragraph{Motivation.} Sarcasm comes in different varieties -- some sarcasms are lucid while some require intense scrutiny of the situation. In such situation, merely the detection of sarcasm without revealing the implicit irony may not be adequate for many use cases. 
For various applications, ranging from feedback analysis in e-commerce to sensitive social media analysis, understanding \textit{why something is sarcastic} is as crucial as detecting negative sentiment in the form of sarcasm. This suffices the requirement explaining the intended sarcasm for every sarcastic posts. To this end, we propose a novel problem -- {\bf Multimodal Sarcasm Explanation} ({\bf MuSE}). The task takes a multimodal (image and its caption) sarcastic post as input, and aims to generate a natural language sentence to explain the intended irony in the sarcastic post. Figure \ref{fig:example} shows two instances of the MuSE task. In the first case, the image shows that a car is parked in front of a building with the user-written caption `\textit{This guy gets a gold star in parking.}'. Taking the inter-modal incongruity into account, we can realize that the user is highlighting the improper car parking as it partially covers the reserved parking slot for handicap person. As an outcome of the MuSE task, we expect to generate a similar explanation for the sarcastic post. Similarly, we show another instance of MuSE in Figure \ref{fig:example:similarity}. Moreover, the instance in Figure \ref{fig:example:difference} highlights the importance of multimodal content for sarcasm detection. Evidently, it is extremely non-trivial to classify the post as sarcastic without image. On the other hand, the instance in Figure \ref{fig:example:similarity} defines both inter- and intra-modal incongruities. The inter-modal incongruity exists between {\em 70mbps} (text) and {\em 0.21mbps} (image), while the intra-modal incongruity is apparent in caption because of the positive words (awesome, love) and negative word (\#wtf).   

\paragraph{Formulation of MuSE.} The task of MuSE is different from the traditional explainable systems that use attention heatmaps \cite{guo2019visual, yao-wan-2020-multimodal-transformer} or similar mechanisms (e.g., SHAP \cite{parsa2020toward}, LIME \cite{pramanick-etal-2021-detecting,mahajan2021explainable}, etc.) to explain the model behavior. In contrast, we project the sarcasm explanation as a natural language generation task. Thus, MuSE's output requires to be a cohesive and coherent English sentence. We formally define MuSE as follows: For a given multimodal sarcastic post $P~=~\langle I,~T[t_1, t_2, ..., t_N]\rangle$, where $I$ and $T[]$ denote the image and caption, respectively, and $t_i$ is the token in the caption, we aim to reveal the intended irony by generating a natural language explanation $E[e_1, e_2, ..., e_D]$, where $\forall t_i, e_j \in Vocab^{English}$ and $e_j$ indicates the token in the explanation.  

\paragraph{New Dataset and Baselines.} To address MuSE, we curate \name, a novel multimodal sarcasm explanation dataset, consisting of $3510$ sarcastic posts with natural language explanations manually generated by expert annotators. To benchmark \name, we design a Transformer based encoder-decoder model. We employ two encoders --one each for text and image-- to obtain the modality-wise latent representations, which is followed by the incorporation of a cross-modal attention module. Finally, a BART-based decoder is added in the pipeline for the explanation generation.

\paragraph{Novelty of MuSE.}
We draw the difference between MuSE and the non-sarcastic interpretation task proposed by \cite{joshi:non-sarcastic:2019} in Figure \ref{fig:example}. The first difference is the incorporation of multimodality in MuSE compared to the text-based non-sarcastic interpretation. The second and the prime difference is that the non-sarcastic interpretations are primarily the negation of the sarcastic texts. In contrast, MuSE is defined to explain the incongruity -- not necessarily with the use of negation. However, we have a few examples for which the explanations can be termed as non-sarcastic interpretations (c.f. Figure \ref{fig:example:similarity}).

\paragraph{Contributions:}
Our main contributions are fourfold:
\begin{itemize}[leftmargin=*]
    \item We introduce MuSE, a {\bf novel task}  aiming to generate a natural language explanation for a given sarcastic post to explain the intended irony. To our knowledge, it is the first attempt at explaining the intended sarcasm.
    \item We develop \name, a {\bf new dataset} consisting of $3510$ triples (image, caption, and explanation) for MuSE.
    \item We benchmark \name\ with a {\bf new Transformer-based encoder-decoder model} which would serve as a strong baseline. Empirical results show its {\bf superiority} over various existing models adopted for this task, across five evaluation metrics.
    \item We perform {\bf extensive human evaluation} to measure the coherence and cohesiveness of the generated explanations by our proposed model.
\end{itemize}

\noindent \textbf{Reproducibility:} The source code and dataset are available at \url{https://github.com/LCS2-IIITD/Multimodal-Sarcasm-Explanation-MuSE}. 

\section{Related Work}
\subsubsection{Sarcasm Detection:} 
Most prior studies focus on detecting sarcasm using one or more modalities. Earlier methods including \cite{7549041} and \cite{felbo-etal-2017-using} use hand-crafted features such as punctuation marks, POS tags, emojis, lexicons, etc., for detecting the sarcastic nature of the input. Recent studies have explored sarcasm detection in multimodal setting. One of the earlier studies on multimodal sarcasm detection  \cite{schifanella2016detecting}. incorporated images along with the corresponding captions for detecting inter-modal incongruity. \citet{hazarika2018cascade} extracted contextual information from the discourse of a discussion thread, encoded stylometric and personality features of the users, and subsequently used content-based features for sarcasm detection in online social media. \citet{cai-etal-2019-multi} exploited the multi-stage hierarchical fusion mechanism for multimodal sarcasm detection. In another work, \citet{oprea2019isarcasm} considered the distinction between the intended and perceived sarcasm and showed the limitations of the existing systems in capturing the intended sarcasm.

\citet{castro-etal-2019-towards} extended multimodal sarcasm detection for conversational dialog systems. The authors introduced a new dataset, MUStARD, for multimodal sarcasm detection. Recently, \citet{bedi2021multi} explored the sarcasm detection task in Hindi-English code-mixed conversations. 

\subsubsection{Sarcastic to Non-Sarcastic Interpretation and Sarcasm Generation:} In addition to sarcasm detection, a few attempts have been made in exploring different aspects of sarcasm analysis. \citet{peled-reichart-2017-sarcasm} and \citet{joshi:non-sarcastic:2019} explored an interesting idea of converting sarcastic text into non-sarcastic interpretation. Both approaches utilize machine translation based systems for generating non-sarcastic interpretation. In contrast, \citet{mishra2019modular} focused on generating sarcastic text given a negative sentiment sentence. All these systems work at the unimodal textual level. In comparison, MuSE incorporates the multimodal signals, aiming to highlight the intended irony of sarcasm instead of converting a sarcastic instance to a non-sarcastic one. 

\paragraph{Natural Language Explanations:} There have been a few studies focusing on explaining model predictions by generating a natural language explanation. \citet{hendricks2016generating} first proposed an explainable model for image classification that targets to explain the predicted label.  \citet{kim2018textual} proposed to  explaining the model actions in self-driving cars. Recently, \citet{kayser2021vil} introduce e-ViL, a benchmark for explainable vision-language (VL) tasks, that establishes a unified evaluation framework and provides the first comprehensive comparison of existing approaches that generate NLEs for VL tasks. A majority of these studies employ NLEs for justifying the outputs of their models. However, in our task, NLE itself is the output of the model which is intended to explain the underlying sarcasm in the given multimodal sarcastic sample. To the best our knowledge, this is the first attempt at generating natural language explanations for multimodal sarcastic posts. 

\begin{figure}[t]
    \centering
    \resizebox{\columnwidth}{!}{
    \begin{tabular}{ccc}
         \includegraphics[height=0.1\textheight]{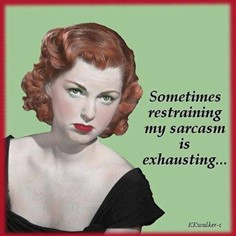} & & \includegraphics[height=0.1\textheight]{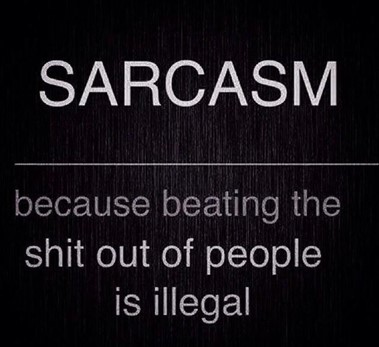}\\
         \multirow{2}{8em}{\textbf{Caption:} \#sarcasm \#sarcastic} & & \multirow{2}{14em}{{\bf Caption:} Some idiots just don 't get it ..  thoughts \#idiot  \#idiotlist.} \\ \\
    \end{tabular}}
    \caption{Posts that are discarded due to explicit sarcasm and do not suffice for an explanation.}
    \label{fig:example:discarded}
\end{figure}

\section{Proposed Dataset}
\label{sec:dataset}
This section elaborates on our effort in developing the \textbf{M}ultim\textbf{O}dal sa\textbf{R}casm \textbf{E}xplanation (MORE) dataset. Since, MuSE demands a sarcastic post, we explore two existing multimodal sarcasm detection datasets -- \cite{schifanella2016detecting} and \cite{sangwan:multimodal:sarcasm:ijcnn:2020} -- to extract the sarcastic posts. \citet{schifanella2016detecting} used hashtag-based approach (\#sarcasm or \#sarcastic) to collect $10000$ sarcastic posts from Twitter, Instagram, and Tumblr. On the other hand, \citet{sangwan:multimodal:sarcasm:ijcnn:2020} manually annotated $1600$ sarcastic posts. Additionally, we explore another multimodal sarcasm detection dataset\footnote{\url{https://github.com/headacheboy/data-of-multimodal-sarcasm-detection}} to collect $10560$ sarcastic posts. In total, we collect $22160$ sarcastic posts. 

Next, we adopt the following annotation guidelines to generate an explanation for each post. 
\begin{itemize}[leftmargin=*]
    \item \textbf{Exclusion:} Following posts are discarded
        \begin{itemize}[leftmargin=*]
            \item Non-sarcastic posts are discarded.
            \item Posts with explicit mention of sarcasm are discarded.
            \item Posts with non-English content are discarded.
            \item Posts that require additional context to interpret sarcasm or the annotators are not familiar with the topics are discarded.
        \end{itemize}
    \item \textbf{Inclusion:}
            Post describing the intra-incongruity (within text, or within image) or inter-incongruity (between image and text) are considered.
    \item \textbf{Annotation Scheme:} Annotators were given the following instructions for generating the explanation.
        \begin{itemize}[leftmargin=*]
            \item All entities including image, caption, hashtags, emojis, etc., are to be considered for interpreting the irony and generating an appropriate explanation. 
            \item In case the underlying sarcasm can be explained in multiple ways, the shorter and simpler explanation is preferred.
            \item Any unrelated topic in explanation is avoided.  
        \end{itemize}    
\end{itemize}

We obtained services of two annotators who carefully examined individual posts in our collection. Following the guidelines, annotators generated explanations of $3510$ sarcastic posts. Out of these samples, \name\ contains $1968$ samples with textual entities as part of the image\footnote{Text written within the image.} along with the captions, while the rest $1542$ samples do not have images and texts overlapped. We call the former \textbf{OCR} samples, while the latter \textbf{non-OCR} samples throughout the paper. The remaining posts are discarded due to one of the aforementioned reasons for exclusion. Two such examples are shown in Figure \ref{fig:example:discarded}.  A brief statistics of the dataset is presented in Table \ref{tab:dataset:stat}. 
\begin{table}[t]
    \centering
    \resizebox{\columnwidth}{!}{
    \begin{tabular}{c|c|c|c|c|c}
         \multirow{2}{*}{\bf Split} & \multirow{2}{*}{\bf \# of Posts} & \multicolumn{2}{c|}{\bf Caption} & \multicolumn{2}{c}{\bf Explanation} \\
         \cline{3-6}
          &  & Avg. length & $|V|$ & Avg. length & $|V|$  \\
         \hline
         
         \hline
         Train & 2983 & 19.75 & 9677 & 15.47 & 5972 \\
         Val & 175 & 18.85 & 1230 & 15.39 & 922 \\
         Test & 352 & 19.43 & 2172 & 15.08 & 1527 \\ \hline
         Total & 3510 & 19.68 & 10865 & 15.43 & 6669 \\ \hline
         
         \hline
    \end{tabular}
    }
    \caption{Statistics of the \name\ dataset.}
    \label{tab:dataset:stat}
    \vspace{-5mm}
\end{table}

\begin{figure}[!t]
\centering
\includegraphics[width=0.75\columnwidth]{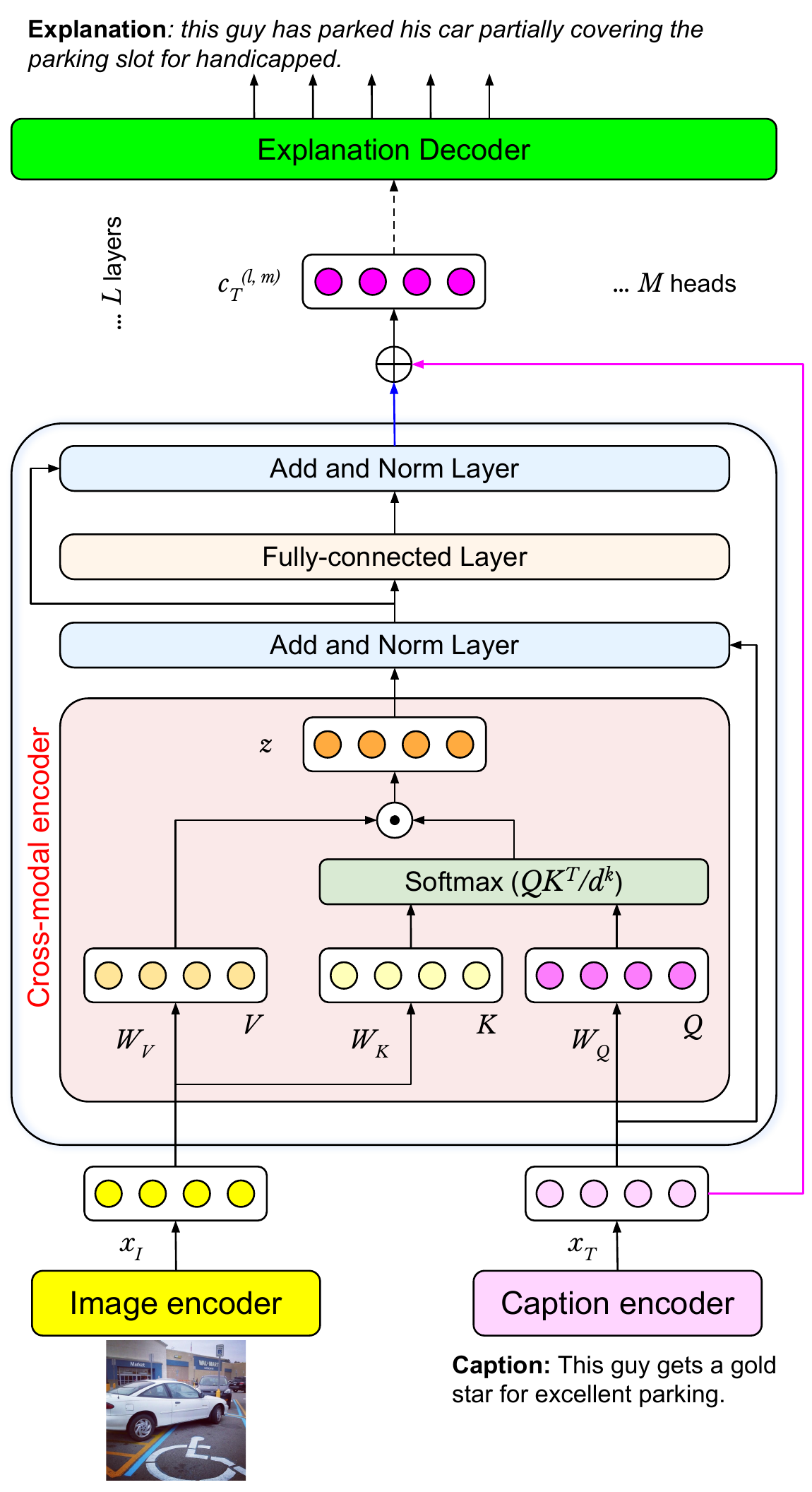}
\caption{Architecture of \modelname.}
\label{fig:model}
\end{figure}

\section{Proposed Benchmark Model}
To generate explanations, we employ \modelname, a multimodal Transformer-based encode-decoder approach\footnote{Note that our aim is not to propose a sophistical model. Rather, we focus on proposing a new task and present a model to benchmark our dataset. Our model is expected to serve as a  strong baseline for MuSE.}. Figure \ref{fig:model} shows the complete architecture of \modelname.

At first, we take both inputs, i.e., images and captions, and pass them through pre-trained VGG-19 \cite{simonyan2014very} and BART \cite{lewis-etal-2020-bart} encoders, respectively. Next, we feed the image ($x_I \in \mathbb{R}^{q \times d^{I}}$) and caption ($x_T \in \mathbb{R}^{r \times d^{T}}$) representations to the multimodal Transformer encoder for cross-modal learning, where $r$ is the number of tokens in caption and $q$ is the number of regions we obtain from the VGG-19 model. Unlike traditional Transformer architecture \cite{vaswani:transformer:2017}, where the same input is projected as `\textit{query}', `\textit{key}', and `\textit{value}', in the multimodal variant, we project the textual representation as `\textit{query}' ($Q \in \mathbb{R}^{r \times d^k}$) and image representation as `\textit{key}' ($K \in \mathbb{R}^{q \times d^k}$) and `\textit{value}' ($V \in \mathbb{R}^{q \times d^k}$). Subsequently, we apply the conventional self-attention mechanism to compute the cross-modal attentive representation $z \in \mathbb{R}^{r \times d^k}$. Taking $d^k = \frac{d^T}{M}$, we incorporate $M=4$ heads for the computation. Following this, we apply layer normalization and fully-connected layers with residual connections to obtain the encoder's output. Finally, we concatenate the textual representation $x$ with the encoder output to obtain the final cross-modal encoder representation $C_T \in \mathbb{R}^{2r \times d^T}$. We feed $C_T$ to the pre-trained auto-regressive BART decoder and fine-tune the entire model on \name\ for the explanation generation. 

\begin{table*}[ht]
    \centering
    \subfloat[All samples\label{tab:exp-results-overall}]{
    \resizebox{0.9\textwidth}{!}
    {
    \begin{tabular}{Hl|c|c|c|c|c|c|c|c|c|c|c|c}
         \hline
         \multirow{2}{*}{\bf Modality} & \multirow{2}{*}{\bf Model} & \multicolumn{4}{c|}{\bf BLEU} & \multicolumn{3}{c|}{\bf Rouge} & {\bf METEOR} & \multicolumn{3}{c|}{\bf BERT-Score} & \bf Sent-BERT \\ \cline{3-9} \cline{11-13}
         &  & B$1$ & B$2$ & B$3$ & B$4$ & R$1$ & R$2$ & R$L$ & & Pre & Rec & F1 & (Cosine)\\ \hline \hline
         \multirow{3}{*}{Text} & Pointer Generator Network
         & 17.54$^{\dagger}$ & 6.31 & 2.33 & 1.67$^{\dagger}$ & 17.35 & 6.90 & 16.00 & 15.06$^{\dagger}$ & 84.8 & 85.1 & 84.9 & 49.42 \\ 
         & Transformer & 11.44 & 4.79 & 1.68 & 0.73 & 17.78 & 5.83 & 15.90 & 9.74 & 83.4 & 84.9 & 84.1 & 52.55 \\ \hline
         \multirow{5}{*}{Text + Image} 
         & MFFG-RNN 
         & 14.16 & 6.10 & 2.31 & 1.12 & 17.47 & 5.53 & 16.21 & 12.31 & 81.5 & 84 & 82.7 & 44.65 \\ 
         & MFFG-Transf 
         & 13.55 & 4.95 & 2.00 & 0.76 & 16.84 & 4.30 & 15.14 & 10.97 & 81.1 & 83.8 & 82.4 & 41.58  \\
         & M-Transf 
         & 14.37 & 6.48$^{\dagger}$ & 2.94$^{\dagger}$ & 1.57 & 20.99$^{\dagger}$ & 6.98$^{\dagger}$ & 18.77$^{\dagger}$ & 12.84 & 86.3$^{\dagger}$ & 86.2$^{\dagger}$ & 86.2$^{\dagger}$ & 53.85$^{\dagger}$ \\ \cline{2-14}
         & \textbf{\modelname\ } & \bf 19.26 & \bf 11.21 & \bf 6.56 & \bf 4.26 & \bf 27.55 & \bf 12.49 & \bf 25.23 & \bf 19.16 & \bf 88.3 & \bf 87.5 & \bf 87.9 & \bf 59.12 \\ \hline
         
    \end{tabular}}}
    \vspace{-2mm}
    
    \subfloat[Non-OCR samples\label{tab:experiment_results_non_ocr}]{
    \resizebox{0.9\textwidth}{!}
    {
    \begin{tabular}{Hl|c|c|c|c|c|c|c|c|c|c|c|c}
         \hline
         \multirow{2}{*}{\bf Modality} & \multirow{2}{*}{\bf Model} & \multicolumn{4}{c|}{\bf BLEU} & \multicolumn{3}{c|}{\bf Rouge} & {\bf METEOR} & \multicolumn{3}{c|}{\bf BERT-Score} & \bf Sent-BERT \\ \cline{3-9} \cline{11-14}
         &  & B$1$ & B$2$ & B$3$ & B$4$ & R$1$ & R$2$ & R$L$ & & Pre & Rec & F1 &  (Cosine) \\ \hline \hline
         \multirow{3}{*}{Text} & Pointer Generator Network & 17.87$^{\dagger}$ & 6.37 & 1.92 & 1.36$^{\dagger}$ & 17.80 & 6.92 & 16.43 & 15.62$^{\dagger}$ & 84.7 & 85.2 & 84.9 & 48.77 \\
         &  Transformer & 11.65 & 5.65 & 1.73 & 0.69 & 17.41 & 6.26 & 16.16 & 10.13 & 83.6 & 85.1 & 84.3 & 48.40 \\
         \hline
         \multirow{3}{*}{Text + Image} & MFFG-RNN & 15.43 & 6.82 & 2.46 & 1.33 & 18.61 & 5.71 & 17.40 & 12.98 & 81.6 & 84.3 & 82.9 & 42.72 \\
         &  MFFG-Transf & 13.28 & 5.35 & 1.49 & 0.26 & 16.80 & 4.35 & 14.90 & 11.19 & 81.3 & 84 & 82.6 & 41.68 \\
         &  M-Transf & 14.91 & 6.90$^{\dagger}$ & 2.66$^{\dagger}$ & 0.83 & 21.05$^{\dagger}$ & 7.08$^{\dagger}$ & 19.34$^{\dagger}$ & 13.91 & 86.5$^{\dagger}$ & 86.3$^{\dagger}$ & 86.4$^{\dagger}$ & 51.77$^{\dagger}$ \\ \cline{2-14}
         & \textbf{\modelname\ } & \bf 19.47 & \bf 11.69 & \bf 6.82 & \bf 4.27 & \bf 27.12 & \bf 12.12 & \bf 24.92 & \bf 19.20 & \bf 88.3 & \bf 87.6 & \bf 88.0 & \bf 56.95 \\ \hline
         
    \end{tabular}}}
    \vspace{-2mm}
    
    \subfloat[OCR samples\label{tab:experiment_results_ocr}]{
    \resizebox{0.9\textwidth}{!}
    {
    \begin{tabular}{Hl|c|c|c|c|c|c|c|c|c|c|c|c}
        \hline
        \multirow{2}{*}{\bf Modality} & \multirow{2}{*}{\bf Model} & \multicolumn{4}{c|}{\bf BLEU} & \multicolumn{3}{c|}{\bf Rouge} & {\bf METEOR} & \multicolumn{3}{c|}{\bf BERT-Score} & \bf Sent-BERT \\ 
         \cline{3-9} \cline{11-14}
         & & B$1$ & B$2$ & B$3$ & B$4$ & R$1$ & R$2$ & R$L$ & & Pre & Rec & F1 &  (Cosine) \\ \hline \hline
         \multirow{3}{*}{Text} & Pointer Generator Network & 17.19$^{\dagger}$ & 6.08 & 2.49 & 1.79 & 16.92 & 6.76 & 15.55 & 14.64$^{\dagger}$ & 84.9 & 84.9 & 84.9 & 49.53 \\
         & Transformer & 10.68 & 4.01 & 1.49 & 0.71 & 17.25 & 5.32 & 15.04 & 8.99 & 83.2 & 84.7 & 83.9 & 53.94 \\
         \hline
         \multirow{3}{*}{Text + Image} & MFFG-RNN & 12.18 & 4.92 & 1.73 & 0.88 & 15.18 & 4.56 & 14.01 & 10.64 & 81.2 & 83.7 & 82.4 & 45.91 \\
         & MFFG-Transf & 12.87 & 4.12 & 1.69 & 0.62 & 15.54 & 3.53 & 14.20 & 9.70 & 81 & 83.6 & 82.3 & 41.13 \\
         & M-Transf & 14.06 & 6.25$^{\dagger}$ & 3.22$^{\dagger}$ & 2.28$^{\dagger}$ & 21.04$^{\dagger}$ & 7.01$^{\dagger}$ & 18.42$^{\dagger}$ & 12.06 & 86.2$^{\dagger}$ & 86.1$^{\dagger}$ & 86.1$^{\dagger}$ & 55.66$^{\dagger}$ \\ \cline{2-14}
         & \textbf{\modelname\ } & \bf 19.40 & \bf 11.31 & \bf 6.83 & \bf 4.76 & \bf 28.02 & \bf 13.10 & \bf 25.66 & \bf 19.55 & \bf 88.2 & \bf 87.5 & \bf 87.9 & \bf 60.82 \\  \hline
         
    \end{tabular}}
    }
    \vspace{-3mm}
    \caption{Comparative analysis on the \name\ dataset. Dagger ($^{\dagger}$) represents the best baseline.}
    \label{tab:exp_results}
    \vspace{-5mm}
\end{table*}

\section{Experiments, Results, and Analysis}
We evaluate the generated explanations both quantitatively (using standard text generation evaluation metrics) and qualitatively (human evaluation). We also furnish comparative analyses at both levels.

\subsection{Comparative Systems}
Due to multimodal nature of the input, we study MuSE for both unimodal (text) and multimodal (text + image) inputs. Therefore, we also employ comparative systems for both the modalities. For text-based baselines, we employ \textbf{Transformer} \cite{vaswani:transformer:2017} and \textbf{Pointer Generator Network} \cite{see-etal-2017-get} for generating explanations. In the multimodal setup, we adopt \textbf{MFFG}, the video summarization system proposed by  \citet{liu-etal-2020-multistage}. The MFFG architecture is a multi-stage fusion mechanism with a forget fusion gate acting as a multimodal noise filter. We compare with both RNN and Transformer variants of MFFG. We also utilize the multimodal Transformer (\textbf{M-Transf})  \cite{yao-wan-2020-multimodal-transformer} originally proposed for machine translation. M-transf and \modelname\ differ in the way they consume multimodal inputs in their encoders. M-transf considers the concatenation of text and image representations for \textit{query} and text representation for \textit{key} and \textit{value}. On contrary, \modelname\ considers text representation for \textit{query} and image representation for \textit{key} and \textit{value} projections. 

\subsection{Experimental Setup}
We perform experiments on \name\ and use 85:5:10 split to create train  ($2983$), validation ($175$), and test ($352$) sets. We employ \textbf{BLEU} (B1, B2, B3, and B4), \textbf{ROUGE} (R1, R2, and R\textit{L}), \textbf{METEOR}, \textbf{BERTScore} \cite{Zhang*2020BERTScore:}, and \textbf{SentBERT} (a BERT-based cosine similarity at the explanation level), as evaluation metrics. SentBERT estimates the semantic closeness between the reference and generated explanations in the Sentence-BERT \cite{reimers-gurevych-2019-sentence} embedding space.

\subsubsection{Task-based pre-training:} 
Since \name\ has limited number of training samples, we utilize the existing multimodal sarcasm detection datasets (c.f. Section \ref{sec:dataset}) to pre-train \modelname's encoder. The pre-training was performed as a binary (\textit{sarcastic} or \textit{non-sarcastic}) classification task. This enables the encoder to learn the distinguishing features for sarcastic posts that can be leveraged in the MuSE task. Subsequently, the pre-trained \modelname\ encoder is then used to train and fine-tune on \name.

\subsubsection{Hyperparameters:} We employ BART \cite{lewis-etal-2020-bart} tokenizer with maximum token length as $256$. We use {\tt AdamW} \cite{DBLP:journals/corr/abs-1711-05101} optimizer with learning rate of $1e-5$ for the single cross-modal encoder and $3e-4$ for the LM head of decoder. We train \modelname\ for $125$ epochs with {\tt batch\_size} $= 16$. During training, the {\tt cross-entropy} loss is monitored over the validation set with image encoder in a frozen state.

\subsection{Experimental Results}
Table \ref{tab:exp_results} shows the comparative results on \name. We perform evaluation\footnote{We train on the complete dataset and evaluate according to the data type, i.e., non-OCR, OCR, or complete.} for three cases -- (a) on complete dataset (both \textit{non-OCR} + \textit{OCR} samples) (Table \ref{tab:exp-results-overall}), (b) only on \textit{Non-OCR} samples (Table \ref{tab:experiment_results_non_ocr}), and (c) only on \textit{OCR} samples (Table \ref{tab:experiment_results_non_ocr}).

In the overall case (c.f. Table \ref{tab:exp-results-overall}), among all the competing methods, M-transf reports the best performance in each case -- albeit the Pointer Generator Network in a few metrics (B1, B4, and METEOR). In comparison, we can observe that \modelname\ outperforms all baselines across five sets of evaluation metrics. We obtain BLEU scores\footnote{Numbers in bracket show improvement points against the best baseline.} of $19.26(+1.72)$, $11.21 (+4.73)$, $6.56  (+3.62)$, and $4.26 (+2.59)$ for B1, B2, B3, and B4, respectively. Similarly, we gain $+6.56$, $+5.51$, and $+6.46$ Rouge points at $27.55$, $12.49$, and $25.23$ in R1, R2, and R\textit{L}, respectively. \modelname\ also reports improved performance in METEOR $19.16 (+4.1)$, BERT-Score $87.9 (+1.7)$ and Sent-BERT $59.12 (+5.3)$.

Experimental results for non-OCR and OCR samples are reported in Tables \ref{tab:experiment_results_non_ocr} and \ref{tab:experiment_results_ocr}, respectively. The objective of evaluating separate results for OCR and Non-OCR samples is to analyze the effect of text in images on \modelname's performance. We observe that \modelname\ does not show any sign of bias towards either the OCR samples or the Non-OCR samples -- in both cases, the obtained results across all metrics are comparable and closer to the overall case. Furthermore, we observe that \modelname\ outperforms M-transf in both Non-OCR and OCR cases. 

         
         
         

\begin{table*}[!htp]
    \centering
    \resizebox{0.85\textwidth}{!}
    {
    \begin{tabular}{c|c|c|c|c|c|c|c|c|c|c|c|c}
        \hline
        \multirow{2}{*}{\bf Data} & \multicolumn{4}{c|}{\bf BLEU} & \multicolumn{3}{c|}{\bf Rouge} & {\bf METEOR} & \multicolumn{3}{c|}{\bf BERT-Score} & \bf Sent-BERT \\ \cline{2-8} \cline{10-12}
        & B$1$ & B$2$ & B$3$ & B$4$ & R$1$ & R$2$ & R$L$ & & Pre & Rec & F1 & (Cosine)\\ \hline \hline
         
        Overall & 17.52 & 9.21 & 4.76 & 2.92 & 24.23 & 9.89 & 22.27 & 16.72 & 86.9 & 87.1 & 87 & 59.57 \\
        Non-OCR instances & 17.42 & 9.17 & 4.12 & 1.86 & 23.84 & 9.6 & 21.93 & 17.11 & 86.8 & 87.1 & 86.9 & 56.82 \\
        OCR-instances & 17.81 & 9.47 & 5.56 & 3.91 & 24.86 & 10.27 & 22.94 & 16.62 & 87 & 87.1 & 87 & 61.68 \\ \hline
         
    \end{tabular}}
    \vspace{-2mm}
    \caption{Ablation results for \modelname$_{ocr}$: Experiment with Image + Caption + OCR text.}
    \label{tab:image:description:results}
    \vspace{-2mm}
\end{table*}
\begin{table*}[!ht]
    \centering
    
    \subfloat[Noun\label{tab:noun}]
    {
    \resizebox{\columnwidth}{!}
    {
    \begin{tabular}{l|c|c|c|c|c|c}
    \hline
         {\bf Model} & \multicolumn{2}{c|}{Total} & \multicolumn{2}{c|}{Non-OCR} & \multicolumn{2}{c}{OCR} \\ \cline{2-7}
         & M-Transf & \modelname & M-Transf & \modelname & M-Transf & \modelname \\ \hline
         {\bf Ref count} & \multicolumn{2}{c|}{3.76} & \multicolumn{2}{c|}{3.75} & \multicolumn{2}{c}{3.80} \\ \hline
         {\bf Gen count} & 3.68 & 3.57 & 3.62 & 3.53 & 3.73 & 3.68 \\ \hdashline
         {\bf Difference} & 1.80 & 1.81 & 1.80 & 1.88 & 1.80 & 1.76 \\
         {\bf Overlap} & 0.68 & 1.03 & 0.63 & 1.05 & 0.73 & 1.06 \\ \hdashline
         {\bf Overlap--Syn} & 0.79 & 1.18 & 0.73 & 1.19 & 0.85 & 1.21 \\
         \hline
    \end{tabular}}
    }
    \hspace{1em}
    \subfloat[Verb\label{tab:verb}]
    {
    \resizebox{\columnwidth}{!}
    {
    \begin{tabular}{l|c|c|c|c|c|c}
    \hline
         {\bf Model} & \multicolumn{2}{c|}{Total} & \multicolumn{2}{c|}{Non-OCR} & \multicolumn{2}{c}{OCR} \\ \cline{2-7}
         & M-Transf & \modelname & M-Transf & \modelname & M-Transf & \modelname \\ \hline
         {\bf Ref count} & \multicolumn{2}{c|}{2.78} & \multicolumn{2}{c|}{2.71} & \multicolumn{2}{c}{2.80} \\ \hline
         {\bf Gen count} & 2.67 & 2.41 & 2.51 & 2.38 & 2.81 & 2.46 \\ \hdashline
         {\bf Difference} & 1.24 & 1.18 & 1.16 & 1.15 & 1.32 & 1.17 \\
         {\bf Overlap} & 0.45 & 0.60 & 0.38 & 0.51 & 0.52 & 0.69 \\ \hdashline
         {\bf Overlap--Syn} & 0.60 & 0.77 & 0.52 & 0.72 & 0.67 & 0.81 \\
         \hline
    \end{tabular}}
    }
    \vspace{-2mm}
    
    \subfloat[Adjectives\label{tab:adjective}]
    {
    \resizebox{\columnwidth}{!}
    {
    \begin{tabular}{l|c|c|c|c|c|c}
    \hline
         {\bf Model} & \multicolumn{2}{c|}{Total} & \multicolumn{2}{c|}{Non-OCR} & \multicolumn{2}{c}{OCR} \\ \cline{2-7}
         & M-Transf & \modelname & M-Transf & \modelname & M-Transf & \modelname \\ \hline
         {\bf Ref count} & \multicolumn{2}{c|}{1.32} & \multicolumn{2}{c|}{1.19} & \multicolumn{2}{c}{1.40} \\ \hline
         {\bf Gen count} & 0.86 & 1.04 & 0.80 & 0.91 & 0.91 & 1.14 \\ \hdashline
         {\bf Difference} & 1.04 & 0.91 & 0.91 & 0.85 & 1.13 & 0.95 \\
         {\bf Overlap} & 0.04 & 0.16 & 0.05 & 0.15 & 0.04 & 0.17 \\ \hdashline
         {\bf Overlap--Syn} & 0.04 & 0.18 & 0.05 & 0.16 & 0.04 & 0.20 \\
         \hline
    \end{tabular}}
    } \hspace{1em}
    \subfloat[Adverb\label{tab:adverb}]
    {
    \resizebox{\columnwidth}{!}
    {
    \begin{tabular}{l|c|c|c|c|c|c}
    \hline
         {\bf Model} & \multicolumn{2}{c|}{Total} & \multicolumn{2}{c|}{Non-OCR} & \multicolumn{2}{c}{OCR} \\ \cline{2-7}
         & M-Transf & \modelname & M-Transf & \modelname & M-Transf & \modelname \\ \hline
         {\bf Ref count} & \multicolumn{2}{c|}{1.03} & \multicolumn{2}{c|}{0.96} & \multicolumn{2}{c}{1.09} \\ \hline
         {\bf Gen count} & 0.78 & 0.69 & 0.73 & 0.61 & 0.80 & 0.76 \\ \hdashline
         {\bf Difference} & 0.79 & 0.72 & 0.84 & 0.71 & 0.76 & 0.74 \\
         {\bf Overlap} & 0.27 & 0.24 & 0.22 & 0.18 & 0.31 & 0.28 \\ \hdashline
         {\bf Overlap--Syn} & 0.27 & 0.24 & 0.22 & 0.18 & 0.31 & 0.28 \\
         \hline
    \end{tabular}}
    }
    \vspace{-2mm}
    \caption{POS-based comparison between the reference (Ref) explanation and generated (Gen) explanation for \modelname\ and M-Transf (the best baseline). Numbers show the average count respective to four PoS tags (Noun, Verb, Adjectives, and Adverb). Ref and Gen counts refer to the avg. frequencies. Difference and Overlap are the avg. word counts. Overlap-Syn is the avg. overlap count with the incorporation of synonym words obtained through WordNET. The incorporation of synonyms improves overlap counts; thus suggesting slightly better explanations at the semantic-level as well.}
    \label{tab:linguistic}
    \vspace{-3mm}
\end{table*}

\subsection{Ablation Study} 
We experiment with a variant of \modelname\ that leverage the \textit{ocr} text extracted from image as the third modality input along with the \textit{caption} and \textit{image} (we call it \modelname$_{ocr}$). To handle the tri-modal case, we introduce two parallel cross-modal encoders -- one for \textit{caption} and \textit{image}, and another for \textit{caption} and \textit{ocr} text. Utilizing the two encoders, we obtain two cross-modal encoded representations, $z\_img \in \mathbb{R}^{r \times d^T}$ and $z\_ocr \in \mathbb{R}^{r \times d^T}$. Since some posts may not contain \textit{ocr} text, we introduce a filter/gating mechanism to intelligently fused the encoded representations. To achieve this, we compute the mean across the sequence length dimension of $z\_img$ and $z\_ocr$ representations. The resulting vectors are concatenated and passed through a 2-layered fully-connected network followed by a sigmoid function to produce a weight $\lambda$. The final encoder representation $C_T \in \mathbb{R}^{r \times d^T}$ is obtained as $(\lambda \times z\_img) + z\_ocr$, which is then passed to the pre-trained BART decoder to produce the explanation.

Table \ref{tab:image:description:results} reports the result of the ablation study. Similar to the earlier case, we evaluate \modelname$_{ocr}$ for the overall, only Non-OCR, and only OCR samples. In comparison to \modelname, we obtain inferior results in all three cases. An interesting observation is that \modelname$_{ocr}$ obtains better results for the OCR samples compared to the Non-OCR examples. This could be because \modelname$_{ocr}$ incorporates the OCR text explicitly in the model. However, the performance was on a lower side perhaps due to the inability of the gating mechanism to learn in the absence of sufficient training data points.  

\begin{figure}[t]
    \centering
    \resizebox{\columnwidth}{!}{
    \begin{tabular}{cc}
        & \\
        & {\bf Caption:} This guy gets a gold star for excellent parking. \\ 
        & \\ 
        & \\ 
        \rowcolor{LightCyan} & {\bf Image description:} A white car parked in a parking lot. \\
        \multirow{-6}{*}{\includegraphics[width=.36\columnwidth]{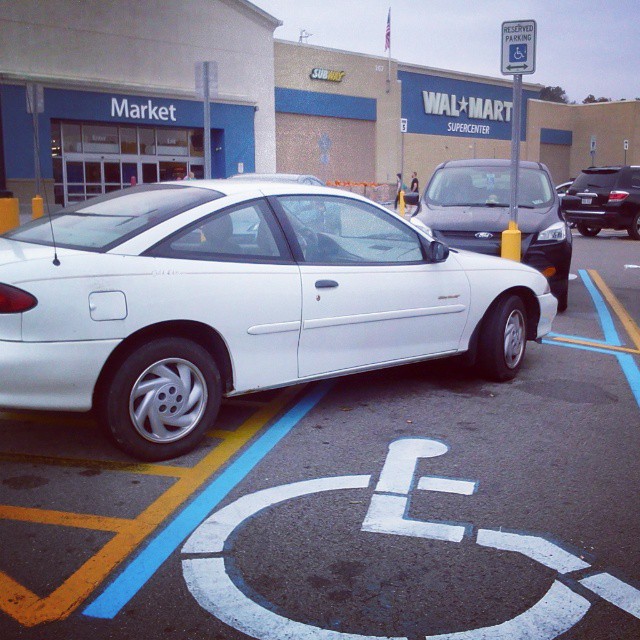}} & \\
    \end{tabular}
    }
    \caption{An example showing that the image description, obtained from Microsoft OSCAR image descriptor, does not capture the caption-specific context that the white car is parked partially covering the parking slot for handicapped. This context provided by the image modality is necessary for understanding the sarcasm.}
    \label{fig:image:description}
    \vspace{-5mm}
\end{figure}

In addition to \modelname$_{ocr}$, we also explore another variant that aims to leverage the image description of an image. The idea seems plausible as extracting desired and relevant information from an image is comparatively non-trivial than extracting the same from text. Therefore, we try to incorporate the image description in place of an image. We generate image description of an image using Microsoft OSCAR \cite{li2020oscar, zhang2021vinvl}, a state-of-the-art image descriptor. However, a manual analysis of the generated image description is found to be non-convincing. Though the image description highlights the key points in an image, features contributing to the irony is missing. One such example is shown in Figure \ref{fig:image:description}. The description `A white car parked in a parking lot.' appropriately describes various spatial features in the image; however, it fails to comprehend the way car is parked -- the target of sarcasm. Therefore, we do not proceed with the image description-based experiments. 

\begin{figure*}[t]
    \centering
    \subfloat[Justify]{
    \resizebox{0.24\textwidth}{!}{
    \begin{tabular}{c}
         \includegraphics[height=0.2\textheight]{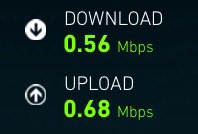}\\
         
         \multirow{2}{20em}{\textbf{Caption:} dear $<$user$>$  please never change . you are perfect in every way .} \\ \\ \\ \hline
         
         \multirow{2}{20em}{\textbf{Ground truth:} the author isn't happy with $<$user$>$'s internet speeds.} \\ \\ \hline
         
         \rowcolor{LightCyan} \\ 
         \rowcolor{LightCyan} \multirow{-2}{20em}{\textbf{Explanation:} the author is pissed at $<$user$>$ for such terrible internet speed.} \\
    \end{tabular}}}
    \subfloat[Weakly Justify]{
    \resizebox{0.24\textwidth}{!}{
    \begin{tabular}{ccccccc}
        \includegraphics[height=0.2\textheight]{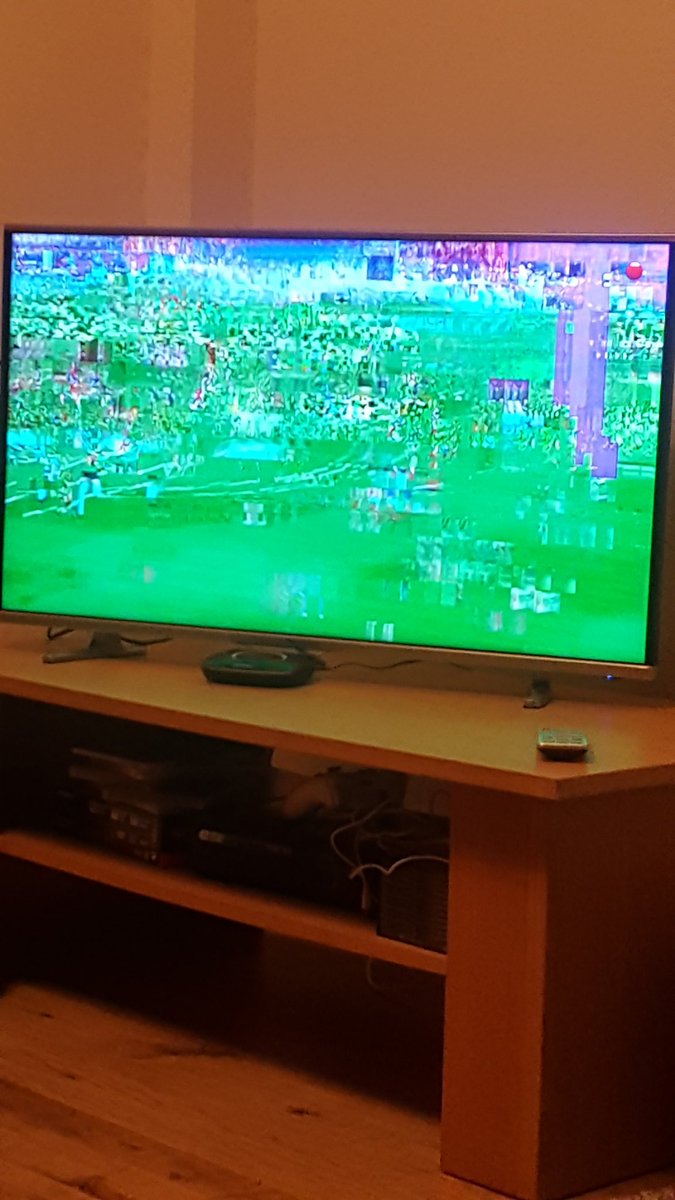}\\
         
        \multirow{2}{20em}{{\bf Caption:} really enjoying the \# atmfcb game tonight $<$user$>$ $<$user$>$} \\ \\ \\ \hline
         
        \multirow{2}{20em}{{\bf Ground truth:} the author can't enjoy the atmfcb game tonight because of such disturbance.} \\ \\ \hline
         
        \rowcolor{LightCyan} \\ 
        \rowcolor{LightCyan} \multirow{-2}{20em}{{\bf Explanation:} the author is pissed at $<$user$>$ for such disturbance.} \\ 
    \end{tabular}}}
    \subfloat[SRI]{
    \resizebox{0.24\textwidth}{!}{
    \begin{tabular}{ccccccc}
        \includegraphics[height=0.2\textheight]{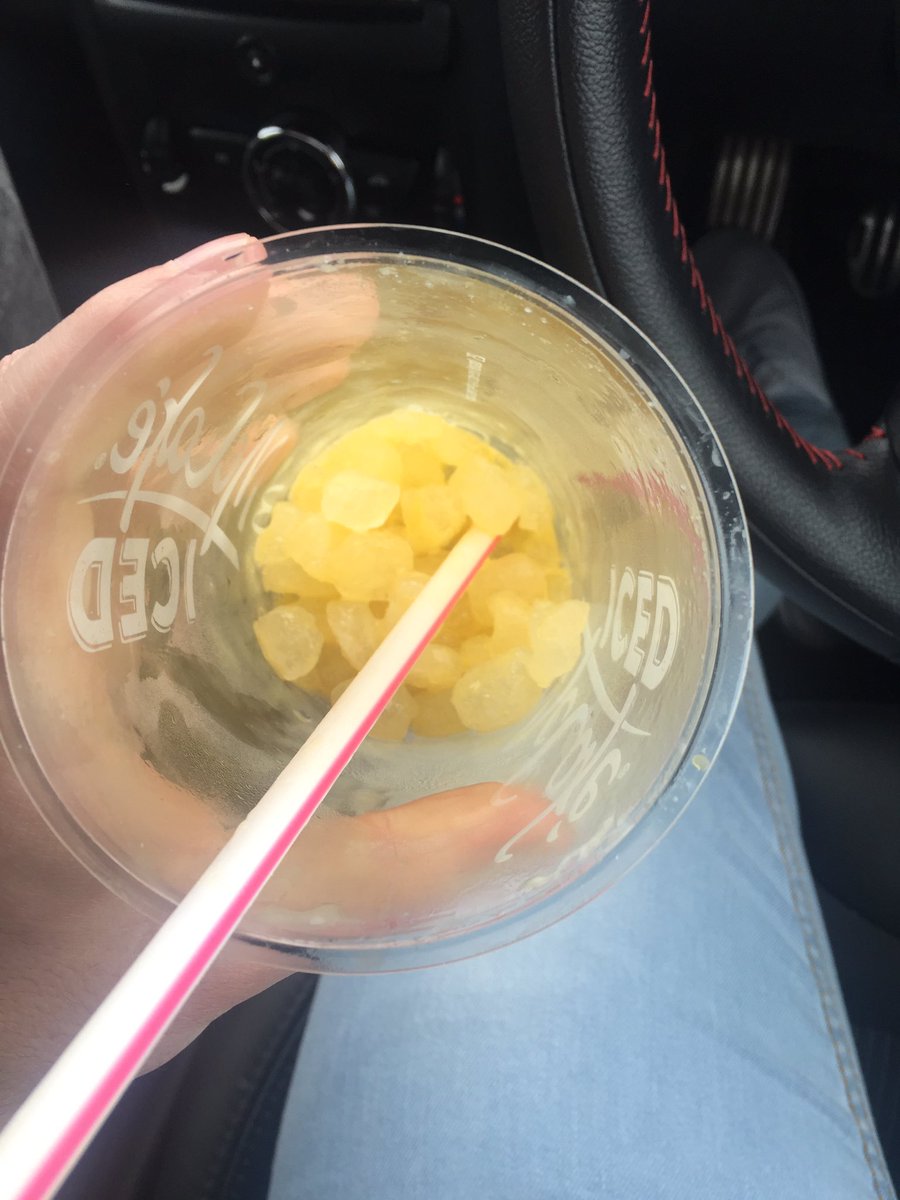} \\
         
        \multirow{2}{20em}{\textbf{Caption:} i just love paying for chunks of ice ... i mean a smoothie $<$user$>$  \# metrocenter \# unacceptable} \\ \\ \\ \hline
         
        \multirow{2}{20em}{\textbf{Ground truth:} the author hates paying for chunks of ice in the name of a smoothie.}\\ \\ \hline
         
        \rowcolor{LightCyan} \\ 
         \rowcolor{LightCyan} \multirow{-2}{20em}{\textbf{Explanation:} the author is pissed at $<$user$>$ for having to wait for a few pieces of ice.}\\
    \end{tabular}}}
    \subfloat[NRI]{
    \resizebox{0.24\textwidth}{!}{
    \begin{tabular}{ccccccc}
        \includegraphics[height=0.2\textheight]{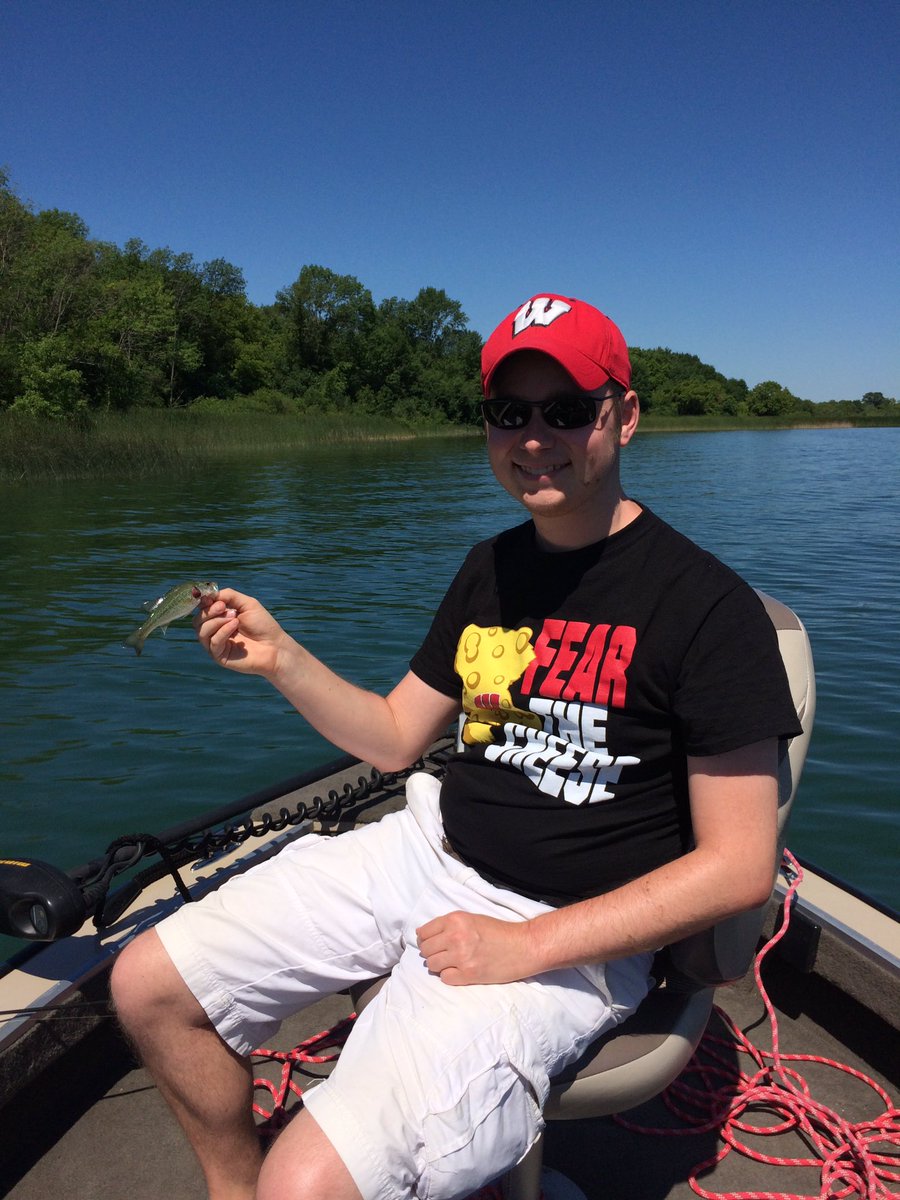}\\
         
        \multirow{2}{20em}{{\bf Caption:} i think i may be the greatest fisherman who has ever lived. just look at the size of my largemouth bass.} \\ \\ \\ \hline
         
        \multirow{2}{20em}{{\bf Ground truth:} the size of the author's largemouth bass is very small.} \\ \\ \hline
         
        \rowcolor{LightCyan} \multirow{1}{20em}{{\bf Explanation:} it's obvious from the stats.} \\
        \rowcolor{LightCyan} \\ 
    \end{tabular}}}
    \vspace{-1mm}
    \caption{Examples of adequacy ratings for the generated explanation by human evaluators. We map the adequacy rating to numeric scale as: \textit{Justify} (1.0); \textit{Weakly Justify} (0.66); \textit{SRI} (0.33); and \textit{NRI} (0.0).}
     \vspace{-2mm}
    \label{fig:adequacy:rating:examples}
\end{figure*}
\begin{table*}[t]
    \centering
    \subfloat[Adequacy and Fluency scores.\label{tab:human_evaluation_adequacy_fluency}]{
    \resizebox{0.28\textwidth}{!}
    {
    \begin{tabular}{c|c|c}
    \multicolumn{1}{c}{} \\
         {\bf Model} & {\bf Adequacy} & {\bf Fluency} \\ \hline
         M-Transf & 0.37 & 0.74 \\
         \modelname\ & 0.69 & 0.89 \\ \hline
    \end{tabular}
    }}
    \hspace{1em}
    \subfloat[Adequacy rating distribution of M-transf and \modelname.\label{tab:human_evaluation_adequacy_rating_distribution}]{
    \resizebox{0.35\textwidth}{!}
    {
    \begin{tabular}{c|c|c|c|c}
         & \multicolumn{4}{c}{\bf Adequacy Rating} \\
         \cline{2-5}
         {\bf Model} & {\bf Justify} & {\bf W. Justify} & {\bf SRI} & {\bf NRI} \\ \hline 
         M-Transf & 15\% & 15\% & 35\% & 35\% \\
         \modelname\ & 65\% & 5\% & 20\% & 10\% \\
         \hline
    \end{tabular}
    }}
    \hspace{1em}
    \subfloat[Fleiss' Kappa scores among 25 human evaluators.\label{tab:fleiss:kappa}]{
    \resizebox{0.28\textwidth}{!}
    {
    \begin{tabular}{c|c|c}
    \multicolumn{1}{c}{} \\
         {\bf Model} & {\bf Adequacy} & {\bf Fluency} \\ \hline
         M-Transf & 0.367 & 0.221 \\
         \modelname\ & 0.401 & 0.175 \\ \hline
    \end{tabular}
    }}
    
    \caption{Human Evaluation: A comparison between \modelname\ and M-Trasnf.} 
    \label{tab:human_evaluation}
    \vspace{-2mm}
\end{table*}

\subsection{Result Analysis} 
\subsubsection{Linguistic view:} In this section, we review the generated explanations from  linguistic aspect. We conduct our analyses according to four content POS tags -- \textit{noun}, \textit{verb}, \textit{adjective}, and \textit{adverb}. Since the POS tags are the carriers of the core semantics of sentences, their comparison would provide us with a sense of their semantic context\footnote{To be clear, we do not claim that POS-based comparison would allow us to compare at the semantic level; instead, it would provide us with a high-level semantic context.}. 
For each case, we count the frequency of respective tag in the generated (Gen) and reference (Ref) explanations. Moreover, to check the closeness of two explanations considering the underlying PoS tag, we also compute the difference and overlap word counts. Furthermore, we also incorporate the synonyms\footnote{Utilizing WordNET: \url{https://wordnet.princeton.edu/}} of a word while computing overlaps in the generated and reference explanations. 

Table \ref{tab:linguistic} presents the count-based comparison between \modelname\ and the best baseline, M-Transf (on avg), for the four tags. The numbers show the average count over the test set for the respective metrics. Though the results suggest that there is a significant gap in generating adequate explanations compared to the reference, we observe that the performance of \modelname\ is better than M-Transf in all cases. Another appealing observation is that the overlap count, albeit small, improves (except for the \textit{adverb} case) with the inclusion of synonyms for \textit{noun} (Table \ref{tab:noun}), \textit{verb} (Table \ref{tab:verb}), and \textit{adjective} (Table \ref{tab:adjective}). It suggests that \modelname's explanations are slightly better at the semantic-level as well. 

\subsubsection{Human Evaluation:}
We also perform human evaluation for assessing the quality of the generated explanations. We randomly sample 50 examples from the test set and ask 25 human evaluators\footnote{Evaluators are the experts in linguistics and NLP and their age ranges in 25-40 years.} to rate the generated explanations (\modelname\ and M-Transf) considering their \textit{adequacy} and \textit{fluency}. The former metric measures the goodness of explanation to reveal the underlying sarcasm, whereas the latter represents the coherency of English explanation. Inspired by \citet{kayser2021vil}, for \textit{adequacy}, the human evaluators are provided with four rating options -- \textit{justify}, \textit{weakly justify}, \textit{somewhat related to input} (SRI), and \textit{not related to input} (NRI). \textit{Justify} highlights the high semantic closeness between the generated and reference explanations; whereas  \textit{weakly justify} represents explanations which reveal the semantic incongruence without reasoning out the sarcastic nature. In contrast, both SRI and NRI categorize the instances with no proper explanations, with the difference that, in SRI, the output refers to some entities related to the input (either in image or caption); however, the outputs in NRI are completely unrelated or random. A few examples related to the four classes of adequacy rating are shown in Figure \ref{fig:adequacy:rating:examples}.

Next, we map the four classes onto numeric scale of [0, 1] -- we assign a score of 1.0 to \textit{justify}, 0.66 to \textit{weakly justify}, 0.33 to \textit{SRI}, and 0.0 to \textit{NRI} samples. Evaluators rate fluency of the generated explanation on the continuous scale of [0,1]. Table \ref{tab:human_evaluation} presents the summary of human evaluation in form of adequacy and fluency scores, adequacy rating distribution, and Fleiss' Kappa \cite{fleiss-kappa-1971} scores among 25 evaluators. From Table \ref{tab:human_evaluation_adequacy_fluency}, we observe that the evaluators showed more confidence in the explanation of \modelname\ than M-Transf for both \textit{adequacy} ($0.69$ vs $0.37$) and \textit{fluency} ($0.89$ vs $0.74$) metrics. 

To compute the adequacy rating distribution, we adopt the majority-voting approach across evaluators to select the adequacy class. The results are shown in Table \ref{tab:human_evaluation_adequacy_rating_distribution}. We observe a significant percentage of samples fall under the \textit{justify} or \textit{weakly justify} categories for \modelname. In contrast, most of the samples belong to the SRI and NRI categories for M-Transf. It further strengthens our claim that \modelname\ yields better explanation than all baselines. Furthermore, we compute Fleiss' Kappa to evaluate the agreement among 25 human evaluators. We observe fair agreement \cite{landis-fleiss-kappa-interpretation-1977} among evaluators for adequacy.

\section{Conclusion}
In this paper, we proposed a novel task of multimodal sarcasm explanation (MuSE), aiming to unfold the intended sarcasm in multimedia posts with a caption and an image.  To address the task, we developed a new dataset, \name, containing $3510$ sarcastic posts annotated with reference explanations in natural language (English) sentence. Further, we presented a strong baseline, \modelname, to benchmark the \name\ dataset. Our evaluation showed that \modelname\ outperforms various baselines (adopted for MuSE) across five sets of evaluation metrics. Moreover, we conducted extensive analyses on the generated explanations. The POS tag and synonym-based linguistics analysis showed that \modelname\ produced semantically accurate output than the best baseline. In addition, the human evaluation with fair Fleiss' Kappa agreement among 25 evaluators upheld the quality of \modelname's explanation in form of higher adequacy scores. We believe that MuSE opens a new avenue in the domains of sarcasm analysis and explainability.

\section*{Acknowledgment}
T. Chakraborty would like to acknowledge the support of the Ramanujan Fellowship, and ihub-Anubhuti-iiitd Foundation set up under the NM-ICPS scheme of the Department of Science and Technology, India. M. S. Akhtar and T. Chakraborty thank Infosys Centre for AI at IIIT-Delhi for the valuable support.

\bibliography{anthology,reference}
\end{document}